\documentclass{article}
\usepackage{spconf,amsmath,graphicx}

\usepackage{times}
\usepackage{epsfig}
\usepackage{graphicx}
\usepackage{amsmath}
\usepackage{amssymb}

\usepackage{cuted}
\usepackage{capt-of}

\usepackage{color}
\usepackage{multirow}
\usepackage[table,xcdraw]{xcolor}

\usepackage[final]{pdfpages}
\usepackage{titling}

\predate{} 
\postdate{}
\date{}

\title{Novel View Video Prediction Using a Dual Representation}

\author{Sarah Shiraz, Krishna Regmi, Shruti Vyas, Yogesh S. Rawat, Mubarak Shah
\\ 
Center for Research in Computer Vision (CRCV), University of Central Florida}

\begin{document}

%
\maketitle



\begin{strip}\centering
\includegraphics[width=0.98\textwidth,height=5.5cm]{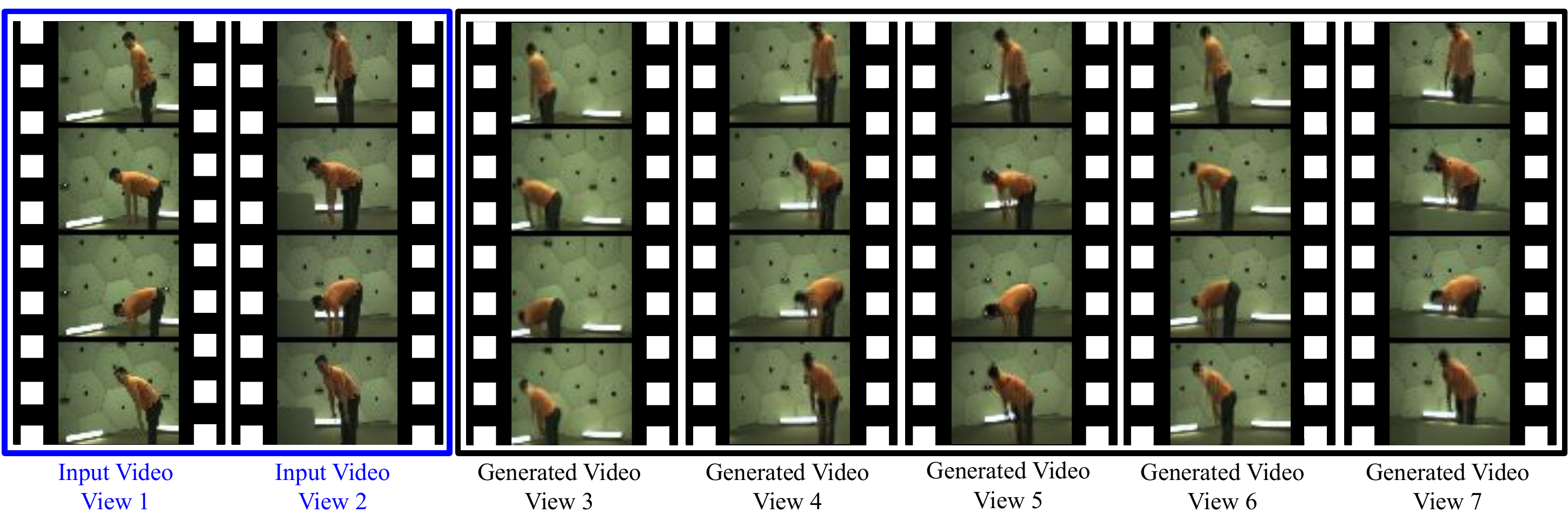}

\captionof{figure}{\textbf{Novel-view Video Prediction.}
Given a set of video clips from single/multiple input viewpoints, the proposed framework can generate a video clip from a novel viewpoint. In the above example, given two input views, our network is able to generate videos from multiple viewpoints, video from five unseen viewpoints are shown here.
}
\label{fig:feature-graphic}

\end{strip}

\begin{abstract}
We address the problem of novel view video prediction; given a set of input video clips from a single/multiple views, our network is able to predict the video from a novel view. The proposed approach does not require any priors and is able to predict the video from wider angular distances, upto 45 degree, as compared to  the recent studies predicting small variations in viewpoint. Moreover, our method relies only on RGB frames to learn a dual representation which is used to generate the video from a novel viewpoint. The dual representation encompasses a  view-dependent and a global representation which incorporates complementary details to enable novel view video prediction. 
We demonstrate the effectiveness of our framework on two real world datasets: NTU-RGB+D and CMU Panoptic. A comparison with the State-of-the-art novel view video prediction methods shows an improvement of 26.1\% in SSIM, 13.6\% in PSNR, and 60\% in FVD scores without using explicit priors from target views.
\end{abstract}


\section{Introduction}

Novel view prediction finds interesting applications in robotics, graphics and augmented reality \cite{seitz1995physically,daribo2010depth,zhou2016view}. Given one or more input views, the task is to predict the seen observations from a novel unseen view. Ample research on novel view prediction has been done for \textit{image based methods} \cite{olszewski2019transformable, sitzmann2019scene}, however novel view prediction for \textit{videos} is much more challenging and relatively unexplored. The image-based approaches, when directly applied to videos, result in temporally incoherent videos as they do not model temporal dynamics.





\begin{figure*}[t]
\begin{center}

\includegraphics[width=0.98\linewidth]{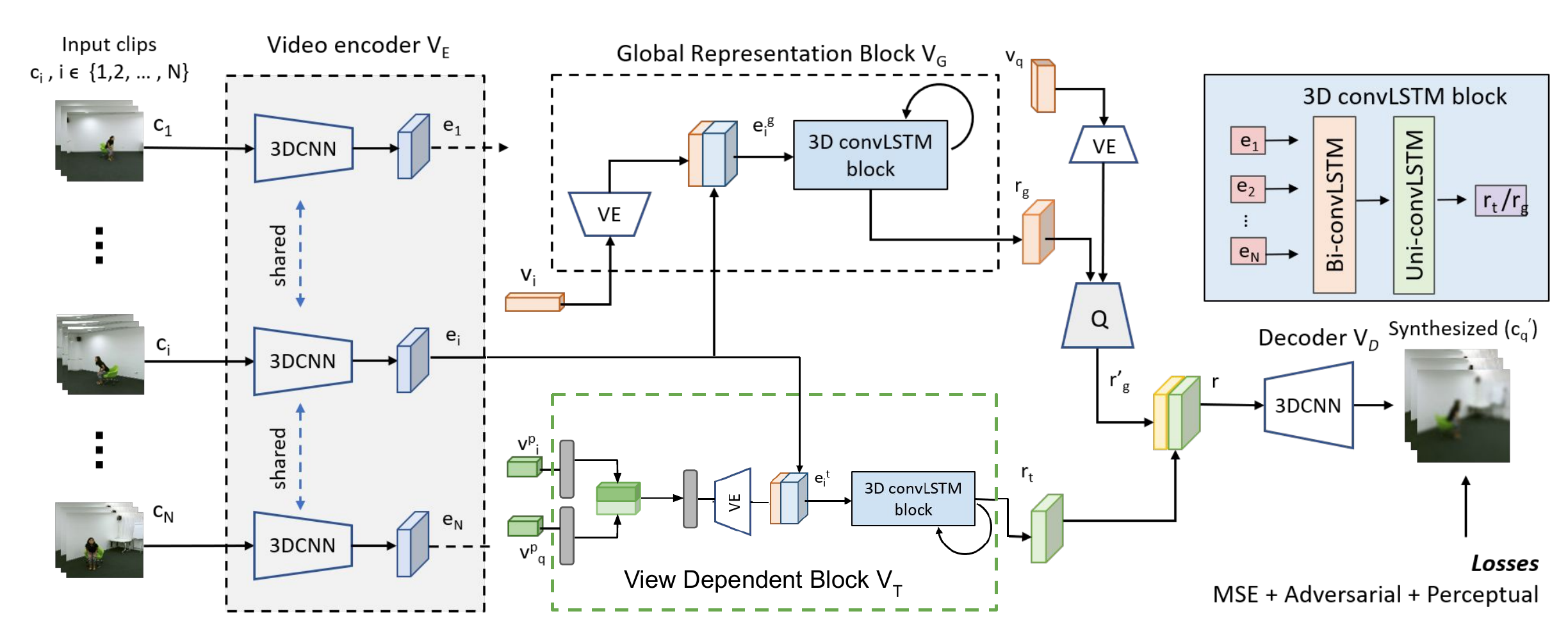}
\end{center}
\caption{\small 
Overall architecture of the proposed network. The Video Encoder $V_E$ takes multiple input clips $c_i$, $i$ $\in$  1,2,.., N from different viewpoints, $v_i$ to learn the encoded features, $e_i$. The Global Representation Block, $V_G$ uses the absolute view parameters of the input, $v_i$. The features $e_i$ are aggregated with view embeddings from view embedder, VE using a 3D convLSTM block. The Query Network, Q retrieves features, $r'_g$ according to the query view $v_{q}$. The view dependent block, $V_T$ on the other hand uses relative view information, using both partial input view $v^p_{i}$ and partial query view $v^p_{q}$, which adds fine view-dependent details to the video. The dual representation, $r$ thus obtained by aggregating the retrieved global representation, $r'_g$ and view dependent representation, $r_t$ is used to synthesizes the query view video, $c_{q}^{'}$.}

  \label{fig:model}
\end{figure*}

In this paper, we address the problem of novel view video prediction without using explicit estimation of 3D structures and geometry based transformations. Most of the video prediction works \cite{vondrick2016generating,tulyakov2018mocogan,palazzo2018generating} in the literature are mainly limited to single viewpoint with the aim of improved dynamics and sharper frames. Only  recently, the authors in  \cite{lakhal2019view} tackled the task of novel view \textit{video} prediction by proposing the use of depth maps and human pose information of the novel view as priors. However, the availability of these priors from the query viewpoint is a big assumption which may not always be valid and requires extra computation.

Considering this limitation, {\em we propose a novel learning based method which generates human action videos from a novel view without any priors and of better video quality}.
Proposed method takes video clips from multiple viewpoints and utilizes a two-stream approach for learning the dual representations: global and view-dependent. The global representation focuses on the general features, such as scene structure, whereas the view-dependent representation focuses on finer details specific to the query view. 
We evaluate our approach with extensive ablations and report state-of-the-art results on two real world datasets, CMU Panoptic \cite{joo2015panoptic} and NTU-RGB+D \cite{shahroudy2016ntu}.
\textbf{Contributions: } \textbf{(1)} We propose a novel-view video prediction framework which can generate videos from unseen query viewpoints. \textbf{(2)} The proposed network integrates both global as well as view-dependent representation for effective novel-view video prediction. 
\textbf{(3)} We provide extensive evaluation of our approach on two real-world datasets, CMU Panoptic \cite{joo2015panoptic} and NTU-RGB+D \cite{shahroudy2016ntu}, achieving significant improvement over the existing methods without using any prior information about the query views. We report 26.1\% improvement in SSIM, 13.6\% improvement in PSNR and 60\% improvement in FVD scores over state-of-the-art viewLSTM \cite{lakhal2019view} on CMU-Panoptic Dataset \cite{joo2015panoptic}.

\section{Related Work}

Early works in multi-view  prediction by Cooke \textit{et al.} \cite{cooke2006multi} and \cite{lombardi2019neural} create a novel view video by exploiting the scene geometry information from multiple sources. 
Similar works by Domański  \textit{et al.} \cite{domanski2009view} discuss novel view video prediction from multiview video sources. 
\cite{lakhal2020novel} estimates the 3D mesh of the person present in the query view and transfer the textures from the 2D images to the mesh.
Recently, Lakhal \cite{lakhal2019view} guided their network to synthesize novel view videos by feeding skeleton and depth priors from the query view. Such priors are strong cues that influence the quality of synthesized videos. 
Learning to synthesize the novel view video without any prior information from the query view is very challenging and our work attempts to solve this problem. We do so by learning holistic  as  well  as  view-dependent representations from multiple input video clips for effective novel-view video prediction.





\section{Method}
Given a set of video clips $c_i$, $i$ $\in$  1,2,.., N from a single or multiple input views, our goal is to synthesize a video clip from a novel viewpoint. We propose a deep neural network which learns a dual internal representation, $r$ by assimilating global and view dependent information required for a novel view generation. Global information is learned in the form of a global representation $r_g$ which encodes the scene as a whole whereas, view-dependent representation $r_t$ helps recover the finer details specific to the query view.
As shown in Figure \ref{fig:model}, input to the proposed network is a set of $ N $ video clips from different view points along with the respective view information, and the output is a video clip from the query view. 
The view information is in the form of a vector, $v$ of dimension $ N \times d $ where d is the number of view parameters for a viewpoint. The $c_{q}^{'}$ is the output clip from the query view. The network can be divided into four parts: a Video Encoder, $V_E$, Representation Learning  Blocks $ V_G $ \& $ V_T $, and a Video Decoder $V_D$. Each part is explained as follow: 





\subsection{Video Encoder $V_E$} 
Video Encoder is a multi-view encoder that takes $N$ video clips separately from multiple views and learns features of each video clip in parallel. We use a 3D CNN based encoder network which is a modified version of an Inflated 3D convNet \cite{carreira2017quo} architecture. The encoder shares weights for all the input video clips. The input to the Encoder is $c_i$ where $i \in  {[1,2,..,N]}$ and the output is the encoded feature $e_{i}$. 

\subsection{Global Representation Block $V_G$}
The goal of this block is to transform the video features $e_{i}$ in a way such that we have a global representation representing the dynamics of the overall scene. To achieve this, video encoding $e_{i}$ are first concatenated with the respective view information, resulting in $e_{i}^{g}$, and then all features are aggregated using a recurrent network. The viewpoint, $v_i$, is a $d$-dimensional vector that contains viewpoint information \textit{i.e.} camera location coordinates, viewpoint angle, principal point offset, focal length etc. It is passed through a View Embedder which performs linear interpolation on it before combining with the video features $e_{i}$. Next,  $e_{i}^{g}, i \in {[1,2,..,N]}$ are aggregated together using a recurrent network consisting of a bidirectional convLSTM followed by a unidirectional convLSTM to learn the global representation $r_g$.

\subsection{View-Dependent Block $V_T$}
The goal of the \textit{View-Dependent Block} $V_T$ is to transform the features of video clips from multiple input viewpoints to the features of a query viewpoint. To do so, it learns the features specific to the query view and help in adding finer details to the synthesized video. To achieve this, we use relative view information \textit{i.e.} partial view vector $v^p_{i}$, comprising of only {viewpoint angles} and {camera location} of the input views $v^p_{i}$ and the query view $v^p_{q}$. The partial viewpoint facilitates the model to focus more on the input view embeddings which are the most similar to query view. Specifically, we apply two fully-connected layers on $v^p_{i}$ and $v^p_{q}$ to learn the relative view features $e_{i}^{t}$, as shown in Figure \ref{fig:model}. 
Similar to aggregation of embeddings in $ V_G $ we use a 3D convLSTM, in particular a bidirectional convLSTM followed by a unidirectional convLSTM to learn the view dependent representation $r_t$. However, the weights of 3D convLSTM block are not shared with $V_G$.


\subsection{Video Decoder $V_D$}
The goal of the Video Decoder, $V_D$, is to integrate the view dependent representation with the global representation to synthesize a video clip from the query view $v_q$. To achieve this, it uses a convoluional-based query network, $ Q $, to extract the query view specific features $r'_g$ from the global representation $r_g$ and then combines it with view dependent representation $r_t$ to get the combined representation $r$. Finally, we use another 3D convNet which uses $r$ and transforms the features to the image space resulting in a synthesized video clip  $c_{q}^{'}$. 

\subsection{Loss Function}
We use a combination of reconstruction, adversarial and perceptual loss as shown in Equation \ref{eqn:total_loss} to train our network.
\begin{equation}
\label{eqn:total_loss}
\mathcal { L } = \lambda _ { r }  { L } _ { r } + \lambda _ { p }  { L } _ { p } + \lambda _ { adv }  { L } _ { adv }
\end{equation}

\noindent where $L _ {r}$, $L _ {p}$ and $L _ {adv}$ are reconstruction, perceptual and adversarial losses respectively. $L _ {p}$ compares the features extracted from conv5\_2 layer of VGG-19 \cite{simonyan2014very} for ground truth and predicted video. We empirically set $\lambda _ {p}$ = $\lambda _ {adv}$ =  0.01 and $\lambda _ {r}$ = 1. 
To improve the realism of the synthesized video clip, we employ a discriminator which \textbf{(a)} distinguishes real video clips from the synthetically generated ones, and \textbf{(b)} it recognizes realistic motion between frames. The loss term is implemented by the minimax loss \cite{goodfellow2014generative}.


  \begin{figure*}[t]
\begin{center}
\includegraphics[width=0.95\linewidth]{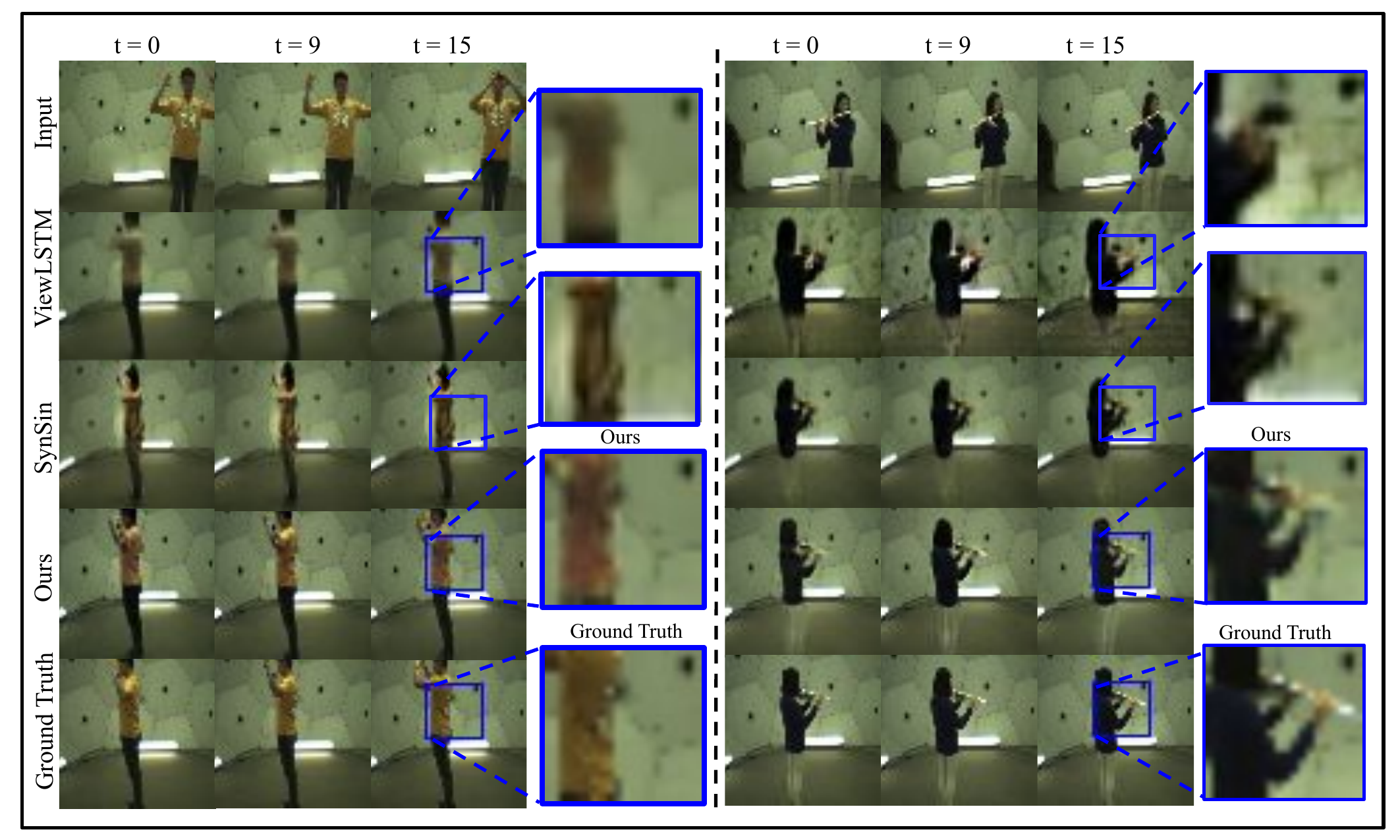}
\end{center}
   \caption{\small Qualitative Results on CMU Panoptic dataset; showing comparison with state-of-the-art method viewLSTM \cite{lakhal2019view} with single-view input
   setup. We show two samples and the resolution is $56 \times 56.$ 
   }
\label{fig:panoptic_3views_comparison}
\end{figure*}

\section{Experiments} 

%
\subsection{Experimental Setup}

\noindent \textbf{Datasets \label{sub:dataset}}
We evaluated our network on following two datasets. \textbf{(1) CMU Panoptic Dataset \cite{joo2015panoptic}} is a large-scale multi-activity real-world dataset with a total of 521 viewpoints.
The cameras are installed in 20 hexagonal panels, with 24 cameras in each panel. The provided viewpoint information vector, $v_i$, consists of camera location coordinates (3D), principal point offset (2D), focal length (2D), distortion coefficients (5D), horizontal-pan (1D) and vertical-pan (1D), therefore $d = 14$.
\textbf{(2) NTU-RGB+D \cite{shahroudy2016ntu}} dataset contains 60 action classes recorded from 3 viewpoints: at horizontal angles of -45 degrees, 0 degree, 45 degrees. We use the cross subject evaluation split of 40,320 and 16,560 clips for training and testing respectively. The viewpoint information vector, $v_i$, for this dataset, consists of  camera height (1D), camera distance (1D), horizontal-pan (1D) and vertical-pan (1D), viewpoint angle (1D), therefore $d = 5$.

\subsection{Novel View Video Prediction: CMU Panoptic} \label{subsection:panoptic_eval}
CMU panoptic \cite{joo2015panoptic} has a large number of views which makes it a lucrative dataset for cross-view video prediction. Considering this, we experiment on a large-scale setting, using a maximum of 72 views for training and testing. We examine the impact of number of training videos on the quality of synthesized videos and compare with SOTA. 

\noindent \textbf{Large-scale Multiview Experiment.} In this setup, we used 72 camera viewpoints. Specifically, we select 3 panels (Number 4, 5 and 17) and use all 24 views from each panel, totalling 72 views.  We  use  56  viewpoints  for  training,  and  16  viewpoints for the testing with a split of 6,244 training and 1,132 test samples.
Training with so many viewpoints can be challenging due to computation cost. So for each iteration we randomly selected, six views from one of the panels, five input views and a query view. Testing followed a similar setup. For each panel, we fix the input viewpoints from view 1 to view 5 $i.e.$ ($v_1$,$v_2$,$v_3$,$v_4$,$v_5$) and test on the remaining 19 viewpoints $i.e.$ the query view points are from $v_6$ to $v_{24}$. 
SSIM \cite{wang2004image} and PSNR for frame quality evaluation
and Frechet Video Distance (FVD) \cite{unterthiner2018towards} to measure the video quality. 

\begin{figure*}[t]
\begin{center}

    \includegraphics[width=0.98\textwidth]{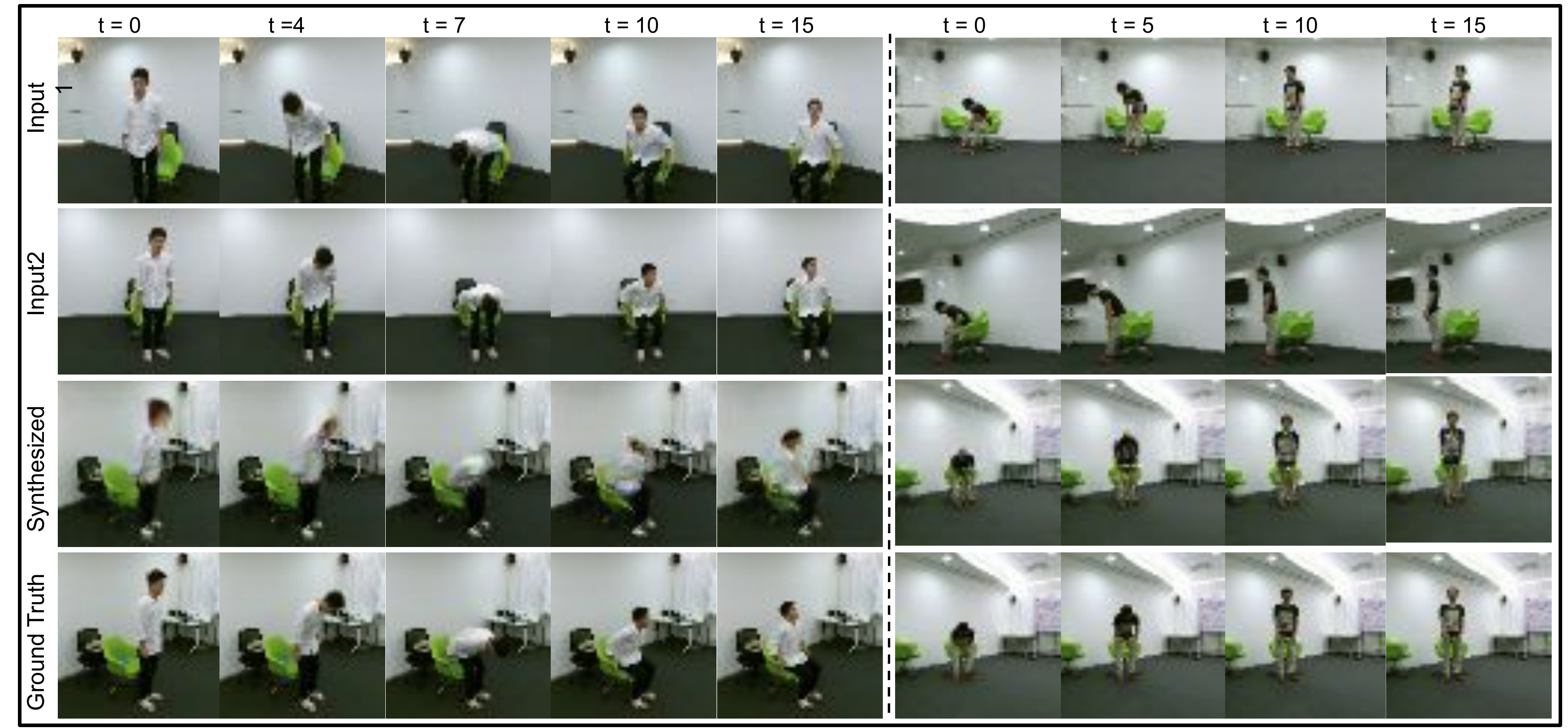}
\end{center}
\caption{Qualitative results on NTU-RGB+D Dataset \cite{shahroudy2016ntu} with multi-input view setup, resolution 56 $\times$ 56. The first two rows show clips from input views $v_{1}$ and $ v_{2}$. The synthesized output from the query view  $v_q$ is shown in the third row followed by the corresponding ground truth frames on the last row. 
}
\label{fig:NTUMultiViewsInput}
\end{figure*}

\begin{table}
 \small
  \centering
  \renewcommand{\tabcolsep}{2mm}  
  
  \caption{\small Quantitative results for Large-scale Multiview experiment (72 views), averaged over 19 query views of each of the three panels of CMU Panoptic at $56 \times 56$ resolution. 
Views ($v_1$,$v_2$,$v_3$,$v_4$,$v_5$) are input views and views $v_6$ to $v_{24}$ are query (output) views.
  }

  \label{tab:72ViewsSSIMResultsTry}
    \begin{tabular}
    {|c|c| c| c|}
                \cline{1-4}
                
        \textbf{ Panels } & \multicolumn{3}{c|} {{\textbf{Metrics}}}
        \\
        \cline{2-4}
         &{\textbf{SSIM}}\textbf{($\uparrow$)} &  {\textbf{PSNR}}\textbf{($\uparrow$)}  & {\textbf{FVD}}\textbf{($\downarrow$)} \\
                \hline
                \hline

        4 &  {0.837} &  {25.15} & {11.52} \\
        5  & {0.848} & {24.9} & {13.42} \\
        17 & {0.845}&  {25.46} &  {13.31} \\
        
          \hline

    \end{tabular}
\end{table}

Table \ref{tab:72ViewsSSIMResultsTry} shows that the proposed method is successful at synthesizing high quality video clips from multiple query views. Owing to the weight sharing nature of the proposed approach, we are able to change the number of testing input clips irrespective of training. Figure \ref{fig:feature-graphic} shows qualitative results with two fixed input views and for different query views. The predicted video frames capture the motion and viewpoints correctly.

\noindent \textbf{Comparison with state-of-the-art}

We compare our results with a video based method viewLSTM \cite{lakhal2019view} which uses skeleton prior from the query view and a frame-based method SynSin \cite{Wiles_2020_CVPR}, both trained from scratch. 
For training, we use three viewpoints, however for a fair comparison with baselines (use one input view), we tested our network employing a single input view setup.
Table \ref{tab:comparison_with_vLSTM_Panoptic_3Views}  shows comparison between these methods in terms of  SSIM, PSNR and FVD scores.  As shown in Table \ref{tab:comparison_with_vLSTM_Panoptic_3Views}, we report 26.1\%, 13.6\%  and 60\% improvement in SSIM, PSNR and FVD scores respectively over viewLSTM \cite{lakhal2019view}, for clips length T=16. Similar improvements are observed when compared with SynSin \cite{Wiles_2020_CVPR} quantitatively.
For visual comparison, some video frames are shown in Figure \ref{fig:panoptic_3views_comparison}. Our model is able to get better quality (sharpness), colors, motion as well as background details. Although, there is ample scope of improvement as compared to ground truth, finer details are much more formed with the proposed approach.


\begin{table}[t]
 \small
  \centering
  \renewcommand{\tabcolsep}{1mm}  
  
  \caption{\small Quantitative comparison of our method with state-of-the-art on CMU Panoptic dataset at $56 \times 56$ resolution for different lengths of synthesized video frames.}  
  \label{tab:comparison_with_vLSTM_Panoptic_3Views}
   \begin{tabular}
    {|c|c|c|c|c|c|c|}
                \hline
         & \multicolumn{6}{c|} {{\textbf{Metrics}}}
        \\
        \cline{2-7}
        \textbf{ Methods }  & \multicolumn{4}{c|} {\textbf{SSIM}\textbf{($\uparrow$)}} &  {\textbf{\small{PSNR}}}\textbf{($\uparrow$)}  & {\textbf{FVD}}\textbf{($\downarrow$)}  
         \\
                \cline{2-7}
         & T=16 & T=24 & T=32 & T=48 & T=16 & T=16
        \\
                \hline
                \hline
        \cite{lakhal2019view} & 0.681 & 0.667 & 0.588 & 0.488 & 22.96 & 22.87  \\
        \cite{Wiles_2020_CVPR} & 0.744 & 0.752 & 0.748 & 0.735 & 21.53 & 17.62   \\
        Ours & \textbf{0.859} & \textbf{0.834} & \textbf{0.809} &  \textbf{0.782}  & \textbf{26.08} & \textbf{9.15}  \\
        \hline
    \end{tabular}
\end{table}


\begin{table}[t]
 \small
  \centering
  \renewcommand{\tabcolsep}{3.5mm}  
  
  \caption{\small SSIM, PSNR and FVD scores for Network Ablations on NTU-RGB+D Dataset \cite{shahroudy2016ntu} 
  using View Dependent (VD) stream, Global Representation (GR) stream and the Full model (Full) on $56 \times 56$ resolution with multi-view input setup. 
  The table shows the average scores for all input-output view combinations.
  }
  \label{tab:network_ablations_2to1}
   \begin{tabular}
    {|c|c| c| c|}
      
                \cline{1-4}

        \textbf{ Networks } & \multicolumn{3}{c|} {{\textbf{Metrics}}}
        \\

        \cline{2-4}
        
         &{\textbf{SSIM}}\textbf{($\uparrow$)} &  {\textbf{PSNR}}\textbf{($\uparrow$)}  & {\textbf{FVD}}\textbf{($\downarrow$)} \\
                \hline
                \hline
VD (tested) &{0.335}&  {12.58} & {18.93} \\
GR (tested) &{0.639}&  {17.45} & {15.68} \\

VD (trained) &{0.563}&  {16.22} & {16.77} \\

GR (trained) &{0.792}&{20.05} &{13.45} \\
Proposed & {\textbf{0.817}} & \textbf{20.77} & \textbf{12.31} \\
          \hline

    \end{tabular}
\end{table}

\subsection{Novel View Video Prediction: NTU-RGB+D}
Since the NTU-RGB+D dataset \cite{shahroudy2016ntu} consists of three camera viewpoints, our experiment uses two views as input and the third view as query. Qualitative results along with the ground truth are visualized in Figure \ref{fig:NTUMultiViewsInput}. Consistent with SSIM and PSNR  values Table \ref{tab:network_ablations_2to1} (Proposed), the network is able to synthesize the persons in the frames as well as the background details of the query viewpoint correctly. We show frames at three different time steps and observe that the synthesized frames are able to capture the motion similar to the ground truth which is also supported by low FVD scores. As ablations studies, we evaluate the role of global and view-dependent blocks separately. First we test the fully trained network by using one block at a time and call them VD(tested) and GR(tested). Secondly, we train both blocks separately and call VD(trained) and GR(trained). The results are shown in Table \ref{tab:network_ablations_2to1} confirming the effect of the two blocks together produces better output.

\section{Conclusion}
We propose a novel approach for unseen view video prediction via a dual representation. The dual representation comprises of global and view dependent representations which enables novel view video prediction. The proposed approach takes single/multiple clips as input for video prediction from the novel view without any priors. Using only RGB videos, our evaluation on two real world datasets shows an improvement over state-of-the-art. 
 
\bibliographystyle{IEEEbib}
\bibliography{strings,refs}

\clearpage








\title{Novel View Video Prediction Using a Dual Representation \\ Supplementary Material}

\name{Sarah Shiraz, Krishna Regmi, Shruti Vyas, Yogesh S. Rawat, Mubarak Shah
}
\address{Center for Research in Computer Vision (CRCV)-University of Central Florida}


%

\maketitle

\section{Additional Network Details}

 \begin{figure*}[t]
 \begin{center}
    \includegraphics[width=.99\textwidth]{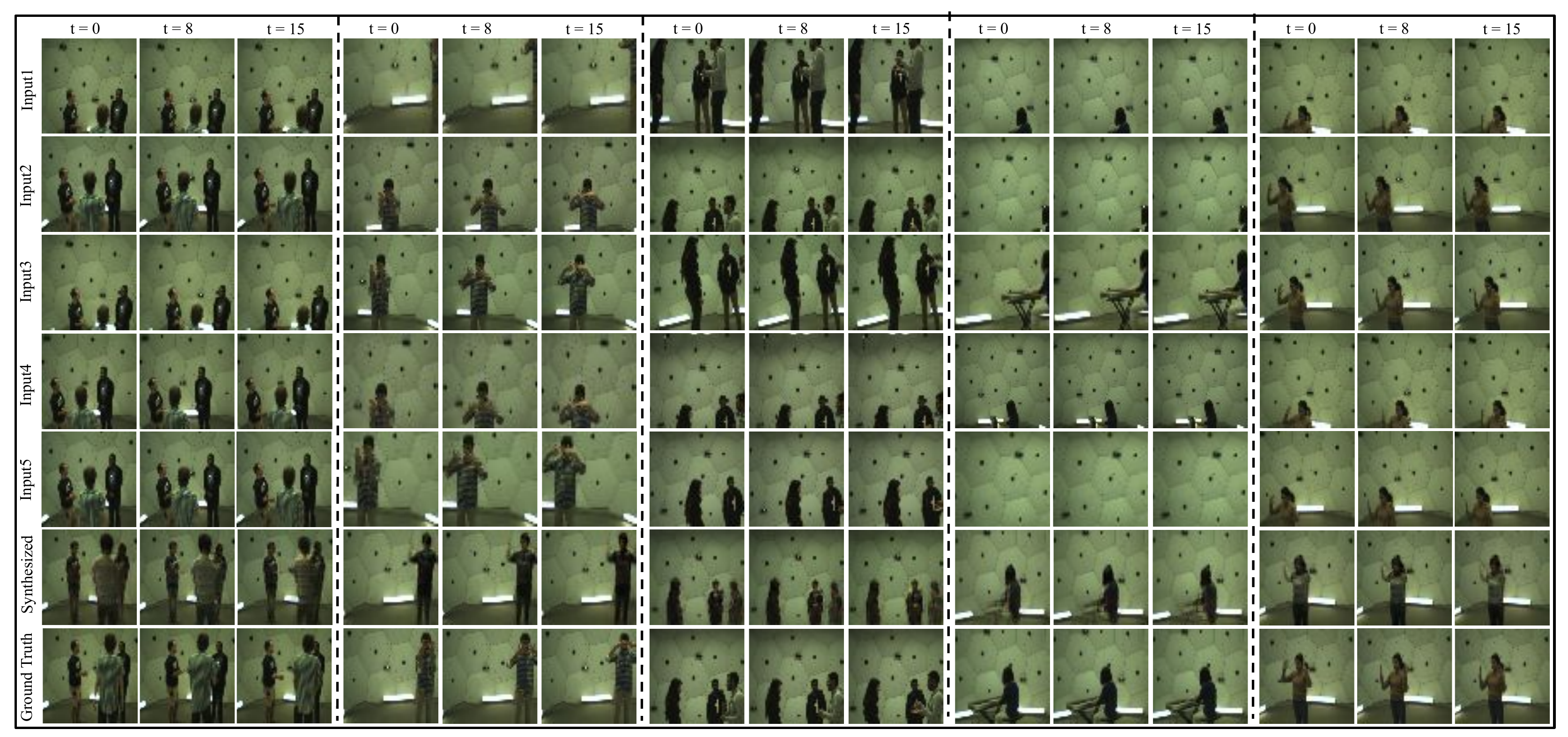}
    \end{center}
\caption{Qualitative results on CMU Panoptic dataset for 72 views experiment with five input views showing three different samples. The resolution is $56 \times 56$.
}
\label{fig:Panoptic72Views}
\end{figure*}

\begin{figure}[t]
\begin{center}

\includegraphics[width=.47\textwidth, height=3cm]{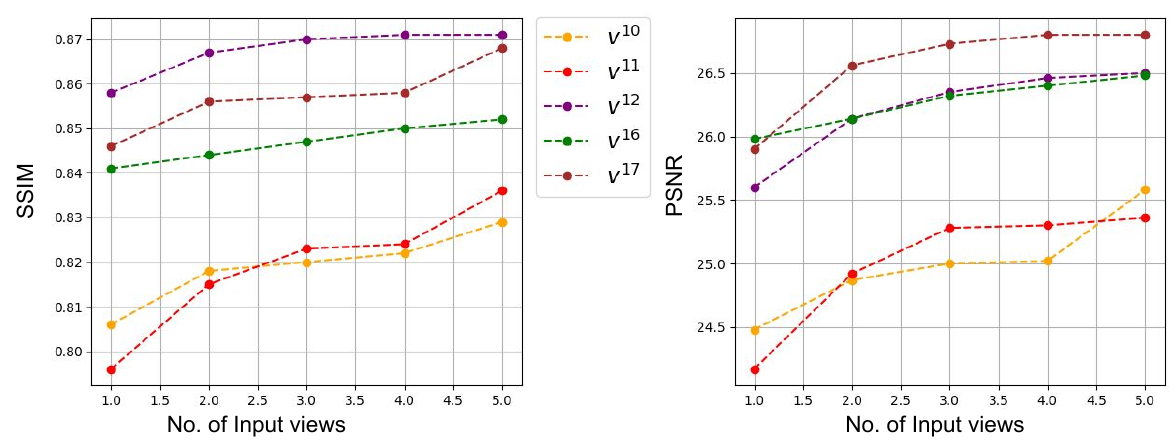}
\end{center}

  \caption{SSIM and PSNR scores vs the number of input views. Higher number of views help better prediction quality. The plots show scores for 5 randomly chosen query views.
  }
\label{fig:effectNumViews}
\end{figure}

\textit{Video encoder} is a modified I3D network\cite{carreira2017quo} where the modifications are: (i) remove max pooling layer after block 2, (ii) for max pooling in block 3 \& 4, stride is set to 2$\times$1$\times$1, (iii) Add 1$\times$1$\times$1 convolutional layer before average pooling, to reduce the number of feature maps to 256. 
Finally, $V_E$ is initialized with I3D weights pre-trained on ImageNet\cite{denglarge}.

\textit{View Embedder, VE} is a trilinear interpolation layer which takes the view information ($v_i$ or $v_q$) of size (1$\times$1$\times d$), where, $d$ is the number of view dimensions.and interpolates the spatial size to match with that of the encoder features in case of the View Dependent Block and the Global Representation Block ( i.e. $v_i$ and $v_i - v_q$) and convLSTM features in the case of Video Decoder ($v_q$) respectively, as we can see from the Network Diagram in the manuscript.

\textit{View Dependent Block, $V_{T}$} consists of two fully-connected layers of size $d^p \times d^p$ and $2d^p \times d^p$ where $d^p$ is the number of view dimensions in $v_{i}^{p} $ or $ v_{q}^{p}$.
Next, it uses a bidirectional convLSTM with 64 3 $\times$ 3 $\times$ 3 kernels; followed by a unidirectional convLSTM with 128 3 $\times$ 3 $\times$ 3 filters in the convolutions.
The output features from the Video Encoder, $e_{i}^{'}$ and the relative view embeddings from the View Embedder are concatenated channelwise and passed as input to this block. 

\textit{Global Representation Block} 
also consists of a bidirectional convLSTM followed by a unidirectional convLSTM. Here, the first convLSTM consists of 128 3 $\times$ 3$ \times$ 3 filters whereas the second one uses 256 3 $\times$ 3$\times$ 3 filters.

\textit{Query Network}
is a single-layer convolutional network with 256 filters and filter size of 5 $\times$ 5 $\times$ 5.

\textit{Video Decoder}
has 8 convolutional layers (combined with upsampling layers) with kernel sizes 5$\times$ 5 $\times$ 5 and 256, 256, 256, 256, 256, 128, 64, 3 filters respectively. The second last layer uses kernel size 3 $\times$ 3 $\times$ 3. The final layer is the Tanh activation.

\textit{Discriminator}
is a 6-layered 3D CNN with 3 $\times$ 3$\times$ 3 kernels and strides of 2 $\times$ 2 $\times$ 2. The number of filters in each layer is [16, 32, 64, 128, 256, 256] respectively. After the last convolution, we add a fully connected layer followed by Sigmoid activation.

\section{Additional Results}





\begin{table*}[t]
 \small
  \centering
  \renewcommand{\tabcolsep}{2mm}  
  
  \caption{\small Quantitative results (SSIM, PSNR and FVD scores) for 72 views experiment on CMU Panoptic dataset on $56 \times 56$ resolution. For each panel, input views are fixed to ($v_1$,$v_2$,$v_3$,$v_4$,$v_5$) and we test output views from $v_6$ to $v_{24}$. 
  }

  \label{tab:72ViewsSSIMResultsTry}
    \begin{tabular}
    {|l|c c c|c c c|c c c|c c c|}

                \cline{1-10}

        \textbf{Query} & \multicolumn{3}{c|} {\textbf{Panel 4}}  & \multicolumn{3}{c|} {\textbf{Panel 5}}& \multicolumn{3}{c|} {\textbf{Panel 17}}
        \\

        \cline{2-10}
        
         \textbf{View}&\textbf{SSIM($\uparrow$)}&  \textbf{PSNR($\uparrow$)} & \textbf{FVD($\downarrow$)}&\textbf{SSIM($\uparrow$)}&\textbf{PSNR($\uparrow$)} &\textbf{FVD($\downarrow$)}&\textbf{SSIM($\uparrow$)}& \textbf{PSNR($\uparrow$)} & \textbf{FVD($\downarrow$)}\\

        \hline
        \hline
        
        \textbf{$v_6$ } &0.814&  24.46 & 7.58&0.852&26.79 &11.79&0.855& 25.94 & 11.80  \\
        \textbf{$v_7$ } &0.795&  24.56 & 11.64&0.843&25.66 &13.17&0.820& 26.06 & 13.17 \\
        \textbf{$v_8$ } &0.794&  24.02 & 10.49&0.842&25.94 &16.01&0.834& 26.15 & 16.04 \\
        \textbf{$v_9$ } &0.813&  24.7 & 10.66&0.827&25.16 &17.65&0.795& 25.18 & 14.23 \\
        \textbf{$v_{10}$ } &0.829&  25.58 & 10.91&0.828&24.99 &14.80&0.807& 25.69 & 14.82 \\
        
        \textbf{$v_{11}$ } &0.836&  25.36 & 11.98&0.823&24.21 &12.72&0.809& 24.33 & 12.67 \\
        \textbf{$v_{12}$ } &0.871&  26.50 & 8.24&0.807&23.17 &10.92&0.784& 24.17 & 10.89 \\
        \textbf{$v_{13}$ } &0.834&  25.09 & 10.80&0.832&23.61 &13.33&0.805& 24.62 & 13.32 \\
        \textbf{$v_{14}$ } &0.811&  24.17 & 15.01&0.808&23.26 &11.86&0.789& 24.25 & 11.85 \\
        \textbf{$v_{15}$ } &0.827&  24.7 & 8.24&0.844&24.43 &16.60&0.797& 23.90 & 16.60 \\
        \textbf{$v_{16}$ } &0.852&  26.48 & 12.69&0.860&25.06 &10.84&0.813& 24.62 & 10.87 \\
        \textbf{$v_{17}$ } &0.868&  26.80 & 11.01&0.835&23.76 &10.50&0.808&24.31 & 10.48 \\
        \textbf{$v_{18}$ } &0.846&  27.43 & 10.44&0.817&24.82 &9.86&0.810& 24.14 & 9.84 \\
        \textbf{$v_{19}$ } &0.829&  25.01 & 12.25&0.836&23.90 &13.71&0.816& 24.60 & 13.70 \\
        \textbf{$v_{20}$ } &0.820&  25.02 & 13.26&0.851&25.68 &13.97&0.827& 24.95 & 13.97 \\
        \textbf{$v_{21}$ } &0.822&  25.06 & 12.14&0.811&24.60 &14.91&0.832& 25.83 & 14.89 \\
        \textbf{$v_{22}$ } &0.810&  24.82 & 10.08&0.839&24.44 &15.04&0.804& 24.65 & 15.04 \\
        \textbf{$v_{23}$ } &0.824&  25.15 & 15.53&0.839&24.73 &14.05&0.811& 24.66 & 14.02 \\
        \textbf{$v_{24}$ } &0.835&  25.61 & 15.94&0.823&23.65 &13.31&0.750& 20.59 & 14.70 \\
                \hline


    \end{tabular}
\end{table*}

\begin{figure*}[t]
\begin{center}
\includegraphics[width=0.78\linewidth]{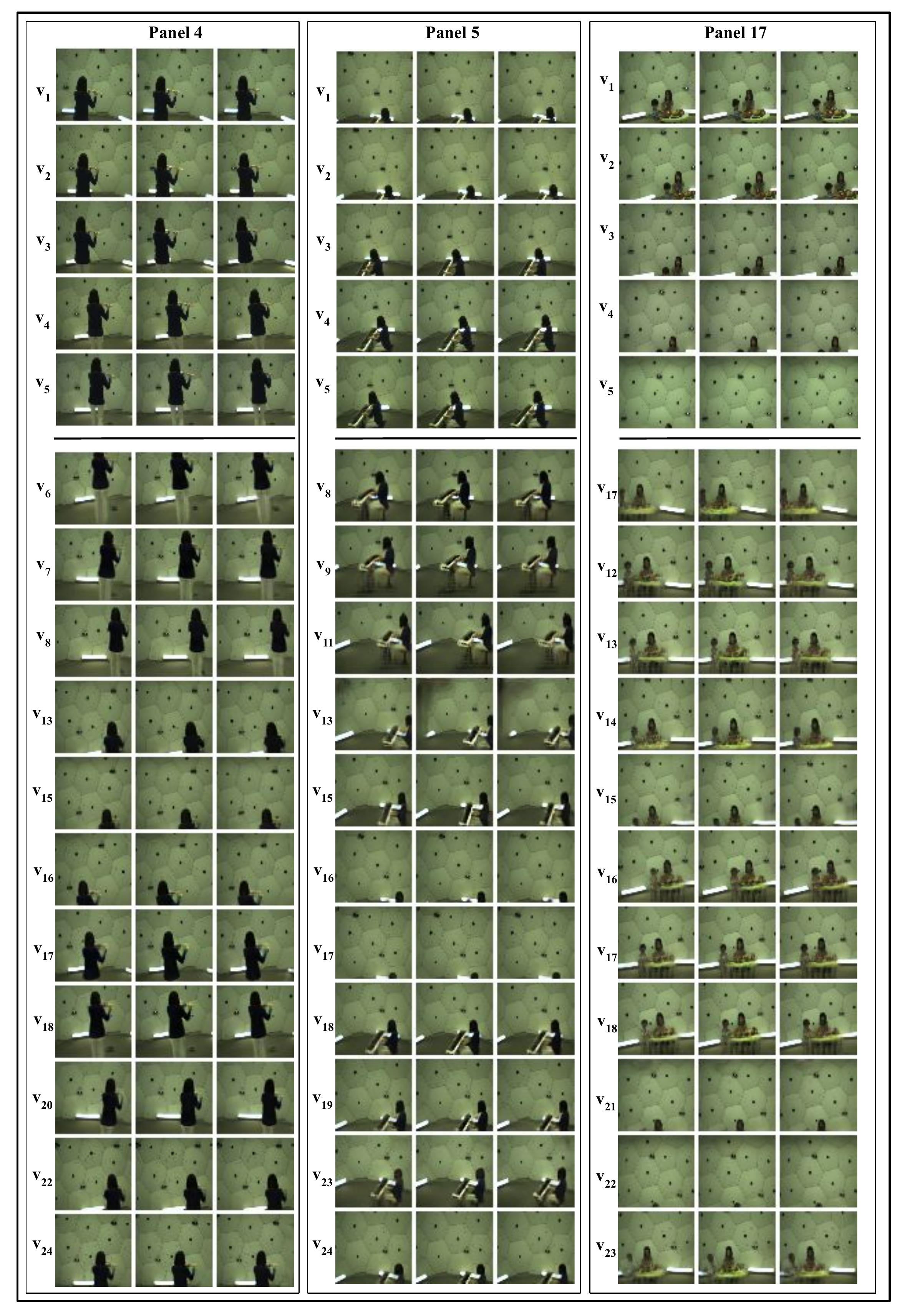}
\end{center}
   \caption{\small  Qualitative Results on CMU Panoptic dataset showing synthesized frames employing  five input views (v$_1$ through v$_5$) and randomly selected query  views from the remaining views ($v_6$ - $v_{24}$) for each panel. 
    We show 3 samples and the resolution is $56 \times 56$. }
\label{fig:panoptic_72Views_panel4}
\end{figure*}

\subsection{CMU Panoptic Dataset \cite{joo2015panoptic}}

\noindent \textbf{Large-scale Multiview Experiment. }
We provide additional qualitative results on CMU Panoptic Dataset. 

Figure \ref{fig:Panoptic72Views} depicts the frames at time t = 0, 8 and 15 for five input views as well as the synthesized frames and the corresponding ground truth frames from the query view.
 
The detailed quantitative evaluation is provided in Table \ref{tab:72ViewsSSIMResultsTry}. The evaluation is conducted per panel separately. For each panel, the  viewpoints from $v_1$ to $v_5$ are fixed as input views and the output video clips for the remaining 19 viewpoints ($v_6$ to $v_{24}$) are synthesized. 
The SSIM, PSNR and FVD scores are provided in Table \ref{tab:72ViewsSSIMResultsTry}. The average scores for panel 4, 5 and 17 are 

For the same configuration, we provide additional qualitative results in Figure \ref{fig:panoptic_72Views_panel4} for randomly selected query views from the pool of viewpoints from $v_6$ to $v_{24}$. It can be observed that video frames and motion are equally good for each of the query views given the five input views as shown.

\begin{table*}[]
 \small
  \centering
  \renewcommand{\tabcolsep}{3mm}  
  
  \caption{\small SSIM and PSNR scores for investigating the effect of varying the number of input views. We randomly select 9 query views $v_k$ from panel 4 and increase the input views from 1 to 5.}
  \label{tab:multiple_views_effect}
    \begin{tabular}{|l|c|c|c|c|c|c|}
    \hline
         & \multicolumn{1}{c|} {\textbf{Query View}}  & \multicolumn{5}{c|} {\textbf{Input Views}}\\
         \cline{2-7}
       \textbf{Metrics}  &
        \textbf{$v_k$} &  \textbf{$(v_1)$} & \textbf{($v_1 ,v_2$)}  & \textbf{($v_1, v_2, v_3$)}  & \textbf{($v_1, v_2, v_3, v_4$)} & \textbf{($v_1, v_2, v_3, v_4, v_5$)}  
        \\
        \hline
        \hline
        & \textbf{$v_6$}  & 0.789 & 0.803 & 0.808 &0.812&0.814 \\
        &  \textbf{$v_{10}$}  & 0.806 & 0.818 & 0.820 & 0.822 &0.829  \\
       &  \textbf{$v_{11}$}  & 0.796 & 0.815 & 0.823 & 0.824 &0.836  \\
        & \textbf{$v_{12}$}  & 0.858 & 0.867 & 0.870 &0.871&0.871 \\
       \textbf{SSIM($\uparrow$)}  & \textbf{$v_{13}$}  & 0.807 & 0.816 & 0.820 &0.819&0.834 \\
        & \textbf{$v_{14}$}  & 0.776 & 0.781 & 0.784 &0.784&0.811 \\
        & \textbf{$v_{16}$}  & 0.841 & 0.844 & 0.847 &0.850&0.852 \\
        & \textbf{$v_{17}$}  & 0.846 & 0.856 & 0.857 &0.858&0.868\\
       & \textbf{$v_{20}$}  & 0.786 & 0.797 & 0.809 &0.812&0.820 \\
    

        
        \hline
        & \textbf{$v_6$}  & 23.51 & 24.06 & 24.27 &24.41&24.46 \\
         &  \textbf{$v_{10}$}  & 24.48 & 24.87 & 25.00 &25.02&25.58 \\
          &  \textbf{$v_{11}$}  & 24.17 & 24.92 & 25.28 &25.30&25.36 \\
        & \textbf{$v_{12}$}  &25.60& 26.14 & 26.35 & 26.46 &26.50 \\
        \textbf{PSNR($\uparrow$)} & \textbf{$v_{13}$}  & 24.82 & 25.16 & 25.25 &25.22&25.09 \\
         & \textbf{$v_{14}$}  &23.34& 23.54 & 23.63 & 23.80 &24.17 \\
         & \textbf{$v_{16}$}  & 25.98 & 26.14 & 26.32 &26.40&26.48 \\
        
        & \textbf{$v_{17}$}  & 25.90 & 26.56 & 26.73 &26.80&26.80 \\
        & \textbf{$v_{20}$}  & 23.99 & 24.29 & 24.64 &24.83&25.02 \\
        
     \hline

    \end{tabular}
\end{table*}


\begin{table*}
 \small
  \centering
  \renewcommand{\tabcolsep}{3mm}  
  
  \caption{\small Quantitative comparison with state-of-the-art viewLSTM\cite{lakhal2019view} on CMU Panoptic dataset at $56 \times 56$ resolution and single-input setup. Our method performs significantly better than SOTA viewLSTM\cite{lakhal2019view} on all three evaluation metrics for all input-output view combinations. 
  }

  \label{tab:comparison_with_vLSTM_Panoptic_3Views}
    \begin{tabular}{|c|c|c|c|c|c|c|c|c|}
        \hline
        \textbf{} & & \multicolumn{6}{c|} {\textbf{Input View $\xrightarrow{}$ Output View}} & \textbf{Average}
        \\
        \cline{3-8}

        \textbf{Method}& \textbf{Metrics} &  \textbf{{$v_1$ $\xrightarrow{}$ $v_2$}} & \textbf{{$v_1$ $\xrightarrow{}$ $v_3$}} & \textbf{{$v_2$ $\xrightarrow{}$ $v_1$}} & \textbf{{$v_2$ $\xrightarrow{}$ $v_3$}} & \textbf{{$v_3$ $\xrightarrow{}$ $v_1$}} &\textbf{{$v_3$ $\xrightarrow{}$ $v_2$}} & \textbf{Score} \\
        \hline
        \hline
        
        & \textbf{{SSIM}}  \textbf{($\uparrow$)}  & 0.675 & 0.684 &0.686 & 0.673 & 0.690 & 0.680 & 0.681\\
      \cite{lakhal2019view} &\textbf{{PSNR}}  \textbf{($\uparrow$)} & 23.03 & 23.16 & 22.56 & 23.07 & 22.69 & 23.24 & 22.96  \\
        & \textbf{{FVD}}  \textbf{($\downarrow$)} & 22.43  &  25.67 & 20.62 & 25.73 & 20.32 & 22.48 & 22.87 \\

                \hline
                
        & \textbf{{SSIM}}  \textbf{($\uparrow$)}  & 0.740 & 0.751 &0.753 & 0.748 & 0.742 & 0.732 & 0.744\\
      \cite{Wiles_2020_CVPR} &\textbf{{PSNR}}  \textbf{($\uparrow$)} & 21.33 & 21.75 & 22.53 & 21.41 & 21.44 & 20.70 & 21.53  \\
        & \textbf{{FVD}}  \textbf{($\downarrow$)} & 18.23  &  18.98 & 16.23 & 17.95 & 17.88 & 16.42 & 17.62 \\

                \hline

        & \textbf{{SSIM}}  \textbf{($\uparrow$)}  & 0.862 &  0.853 & 0.852 & 0.861 & 0.858 & 0.876 & \textbf{0.859}\\
        \textbf{Ours} &\textbf{{PSNR}}  \textbf{($\uparrow$)}  & 26.37 &  25.48 & 25.85 & 25.58 & 26.34 & 26.87 & \textbf{26.08}\\
        & \textbf{{FVD}}  \textbf{($\downarrow$)} & 8.51 & 10.12 & 9.21 & 9.98  & 8.93 & 8.20  & \textbf{9.15} \\

            \hline

    \end{tabular}
\end{table*}


\noindent \textbf{Effect of Multiple Input Views. }


This analysis is conducted to understand the impact of multiple input views on the quality of the synthesized videos. 
The number of input clips is increased from 1 to 5 to synthesize the video from the query view and report the results in Figure \ref{fig:effectNumViews}. As expected, with an increase in the number of input views the video quality improves. This validates that the network is able to capture and aggregate the features pertinent to each view and exploit them to synthesize the query view.

The SSIM and PSNR scores are reported for the synthesized video frames in Table \ref{tab:multiple_views_effect}. The input and query views are selected from Panel 4 for this analysis. We observe that the scores improve as the number of input view increases from single to multiple views. This is possible since the multiple input views are successful at providing greater scene details useful to learn to synthesize the query view.

\begin{figure*}[t]
\begin{center}
\includegraphics[width=0.92\linewidth]{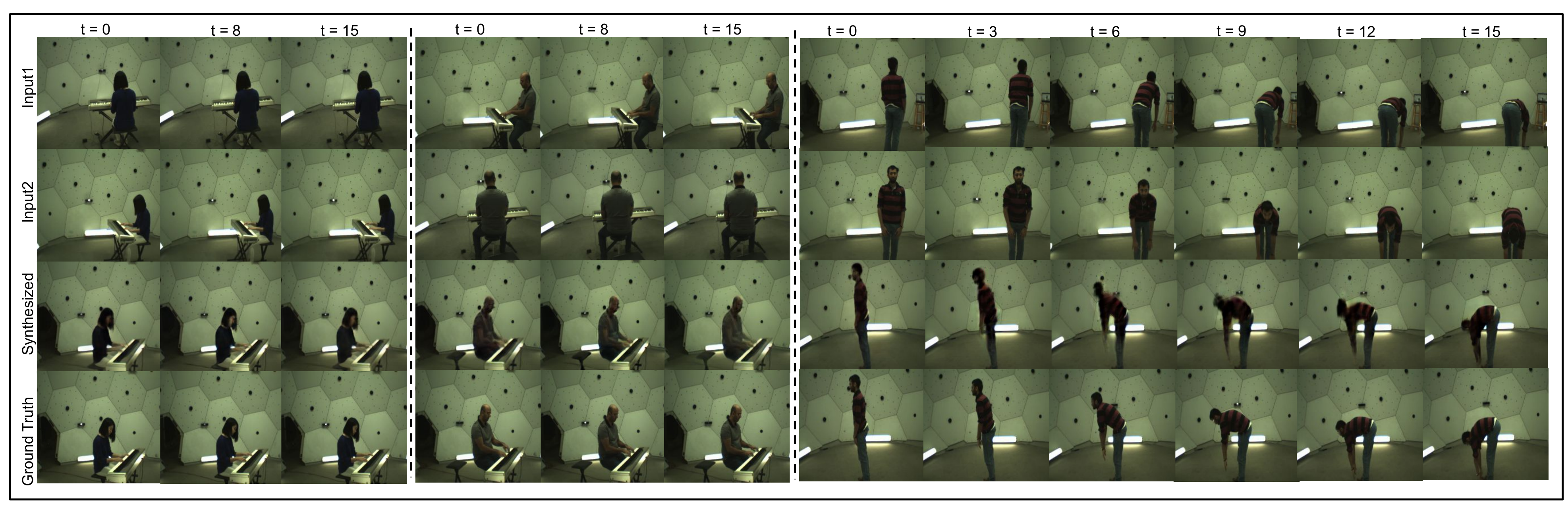}
\end{center}
   \caption{\small Qualitative Results on CMU Panoptic dataset with multi-view input
   setup. We show three samples and the resolution is $224 \times 224.$ 
   }
\label{fig:panoptic_3views_224x224}
\end{figure*}

\noindent \textbf{Higher Resolution Experiment}
For the same setup, we also conduct experiment on 224 $\times$ 224 resolution. The qualitative results are shown in Figure \ref{fig:panoptic_3views_224x224}. The model is able to preserve fine details and produce sharper frames with an SSIM and PSNR score of 0.812 and 24.55, respectively.


\begin{figure*}[t]
\begin{center}
   \includegraphics[width=0.9\linewidth]{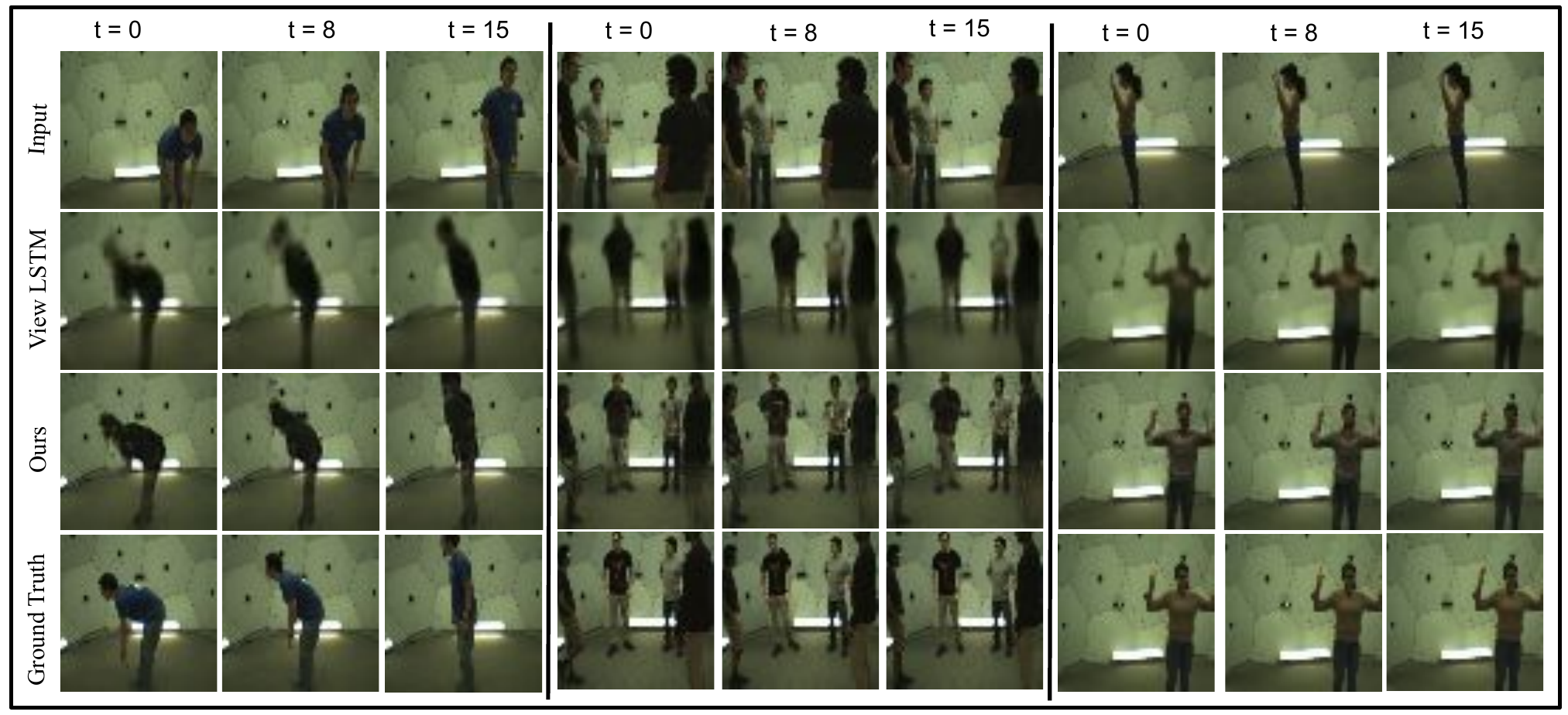}
\end{center}
   \caption{\small Qualitative Results showing the comparison of our proposed method with state-of-the-art method viewLSTM \cite{lakhal2019view} on CMU Panoptic dataset. We show 3 samples, each with frames at time t =0, 8 and 15; and the resolution is $56 \times 56$.}
\label{fig:panoptic_3views_comparison1}
\end{figure*}

\noindent \textbf{Comparison with viewLSTM\cite{lakhal2019view}. }
Here, we provide additional qualitative and quantitative evaluation of our proposed approach with state-of-the-art  viewLSTM \cite{lakhal2019view} in Figure \ref{fig:panoptic_3views_comparison1} and Table \ref{tab:comparison_with_vLSTM_Panoptic_3Views} respectively. The experimental details are provided in the manuscript in Section 5.1.

In Figure  \ref{fig:panoptic_3views_comparison1}, we show three samples from different input viewpoints, each at three time instances, t = 0, 8 and 15. The actors in the images are performing different actions. Qualitatively, our proposed method is successful at synthesizing better images with sharper foreground-background boundaries and also capture the motion of the person along the temporal sequence.

The detailed quantitative result in Table \ref{tab:comparison_with_vLSTM_Panoptic_3Views} provides the performance of our method and the state-of-the-art  viewLSTM \cite{lakhal2019view} for view transformation between the Panels 2,4 and 10 (Node 10). The SOTA method uses the query view skeleton priors to synthesize the query RGB frames, whereas our method doesn't utilize the skeleton information from the query view. We observe that our proposed method outperforms the existing SOTA by huge margins for all evaluation metrics.

\subsection{NTU-RGB+D Dataset \cite{shahroudy2016ntu}}
Refer Section 4.2 in the manuscript for the summarized quantitative analysis on network ablations. Here, we provide the detailed network and loss ablation in terms of quantitative results and qualitative  visualizations on NTU-RGB+D Dataset along with the comparison with existing works.

\begin{figure}
\begin{center}
    \includegraphics[width=0.45\textwidth]{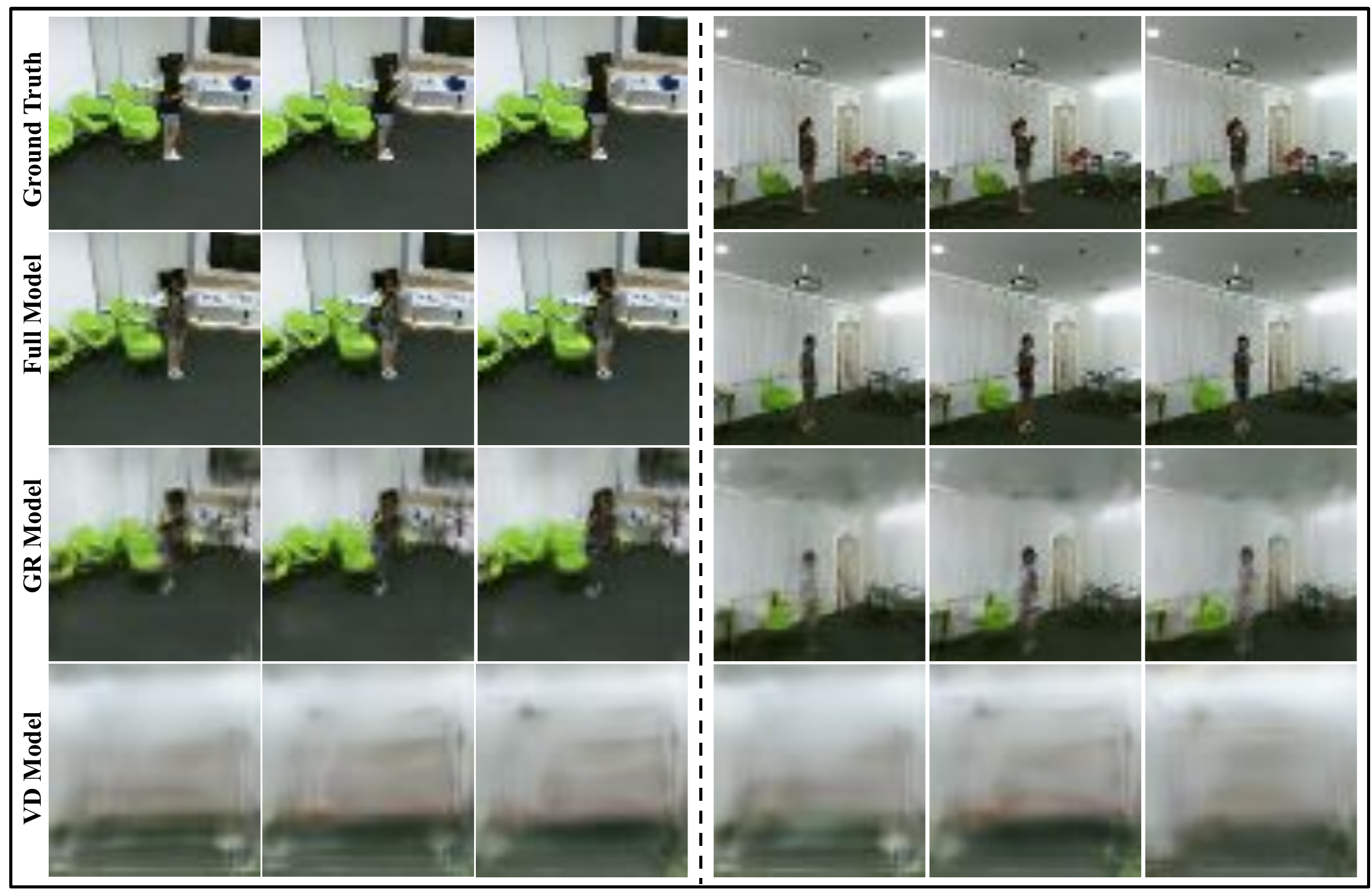}
\end{center}
    
\caption{Qualitative results showing Network Ablations on two samples of NTU-RGB+D Dataset \cite{shahroudy2016ntu}, showing the comparison between View Dependent (tested), Global Representation(tested) and Full Model at $56 \times 56$ resolution for multiple input view setup.
}
\label{fig:NTUNetworkAblationVDGR}
\end{figure}

\begin{figure}[t]
\begin{center}
  \includegraphics[width=.45\textwidth]{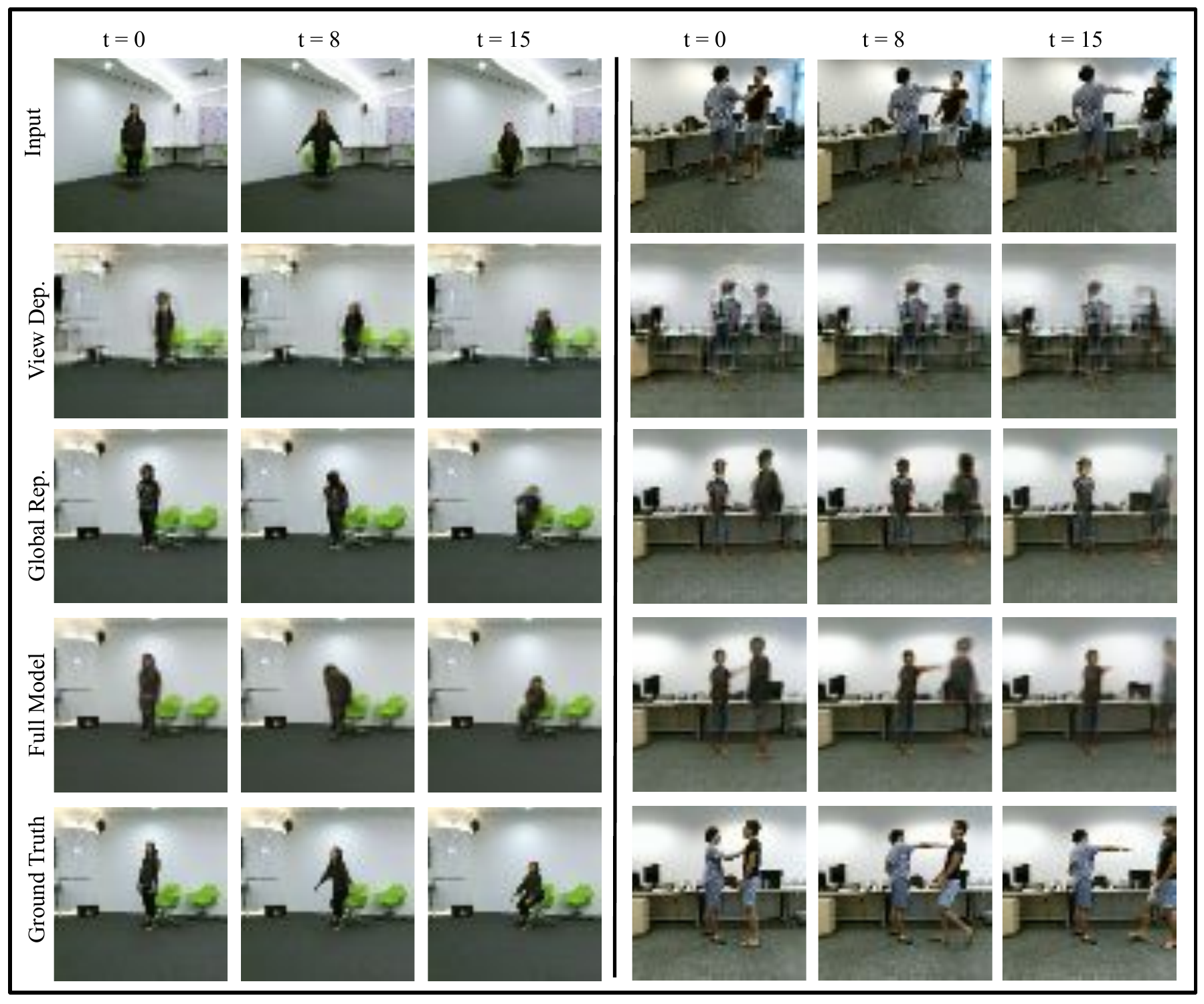}
\end{center}
  \caption{Qualitative results showing Network Ablations on two samples of NTU-RGB+D Dataset \cite{shahroudy2016ntu}, showing the comparison between View Dependent (trained), Global Representation (trained) and Full Model at $56 \times 56$ resolution for single input view setup.
  }
\label{fig:network_ablations}
\end{figure}

\noindent \textbf{Network Ablations.}
We performed two types of ablations for evaluating the role of the global and view-dependent representations. First, we tested the fully trained network by using only one representation at a time. Figure \ref{fig:NTUNetworkAblationVDGR} shows the testing results with global representation $r_g$ and/or view-dependent representation $r_t$. With global representation $r_g$, the foreground and background are almost formed however view-dependent representation $r_t$ is not meaningful alone. The $r_t$ helps by filling in the view specific finer details and only makes sense along with the global representation. It is visibly obvious that best results are obtained with the dual representation $ r $. The detailed quantitative evaluation for single and multi-input setup is shown in Table \ref{tab:network_ablations_2to1} (Row 1-2).

\begin{table*}[]
 \small
  \centering
  \renewcommand{\tabcolsep}{0.4mm}  
  
  \caption{\small SSIM, PSNR and FVD   scores for Network Ablations on NTU-RGB+D Dataset \cite{shahroudy2016ntu} using View Dependent (VD) stream, Global Representation (GR) stream and the full model (Full) on $56 \times 56$ resolution. The left panel shows the results for multi-input view setup, whereas the right panel shows single-input view setup results.
  }

  \label{tab:network_ablations_2to1}
    \begin{tabular}{|c|c|c|c|c|c||c|c|c|c|c|c|c|}
    \hline
        & & \multicolumn{3}{c|}{\textbf{Multi-Input Setup}} & \textbf{Avg} & \multicolumn{6}{c|}{\textbf{Single Input Setup}} & \textbf{Avg}
        \\
        \cline{3-5}  \cline{7-12}
       \textbf{Network}  &
        \textbf{Metric} &  \textbf{($v_1 , v_2$) $\xrightarrow{}$ $v_3$} & \textbf{($v_2 , v_3$) $\xrightarrow{}$ $v_1$}  & \textbf{($v_1 , v_3$) $\xrightarrow{}$ $v_2$} & \textbf{Score}  &  \textbf{$v_1$ $\xrightarrow{}$ $v_2$} & \textbf{$v_1$ $\xrightarrow{}$ $v_3$}  & \textbf{$v_2$ $\xrightarrow{}$ $v_1$} & \textbf{$v_2$ $\xrightarrow{}$ $v_3$} & \textbf{$v_3$ $\xrightarrow{}$ $v_1$} & \textbf{$v_3$ $\xrightarrow{}$ $v_2$} & \textbf{Score}
        \\
        \hline
        \hline
        
        & \textbf{SSIM($\uparrow$)}  & 0.313 & 0.352 & 0.340 &0.335& 0.323 & 0.327 & 0.329 &0.316&0.315&0.323&0.323 \\

     \textbf{VD}\small{(tested)}   & \textbf{PSNR($\uparrow$)}  &12.67 & 13.20 & 11.88 &12.58& 12.09 & 12.52 & 12.59 &11.92&11.89&12.11&12.18\\

      & \textbf{FVD($\downarrow$)}  &18.20 & 18.48 &18.03 &18.23 & 19.23 & 19.93 & 18.96 &18.92&17.78&17.67&18.74 \\

        \hline

       &  \textbf{SSIM($\uparrow$)}  & 0.634 & 0.647 & 0.637 &0.639 & 0.634 & 0.627 & 0.629 &0.624 & 0.628&0.631 & 0.629 \\
  
       \textbf{GR}\small{(tested)}   &  \textbf{PSNR($\uparrow$)}  & 17.23 & 17.72 &17.40&17.45 & 16.12 & 16.96 &16.66 &17.02 & 16.67& 16.31  & 16.62 \\
       
     \textbf{}  &  \textbf{FVD($\downarrow$)}  & 15.45& 15.82 &15.76 & 15.68 & 15.39 & 15.87 & 16.02 & 16.28 & 16.71 &15.39 & 15.94 \\
\hline

        & \textbf{SSIM($\uparrow$)}  & 0.548 & 0.579 & 0.563 &0.563& 0.539 & 0.517 & 0.548 &0.547&0.537&0.519&0.534 \\

     \textbf{VD}\small{(trained)}   & \textbf{PSNR($\uparrow$)}  &16.13 & 16.21 & 16.32 &16.22& 15.86 & 15.62 & 15.59 &15.59&15.63&15.747&15.67\\

      & \textbf{FVD($\downarrow$)}  &16.84 & 16.57 &16.92 &16.77 & 17.48 & 17.37 & 16.77 &17.11&16.90&17.47&17.18 \\

        \hline

       &  \textbf{SSIM($\uparrow$)}  & 0.783 & 0.814 & 0.780 &0.792 & 0.781 & 0.777 & 0.804 &0.774&0.799&0.773 & 0.784 \\
  
       \textbf{GR}\small{(trained)}   &  \textbf{PSNR($\uparrow$)}  & 19.99 & 20.67 &19.50&20.05 & 19.56 & 19.83 &20.2 &19.67&20.04& 19.23  & 19.755 \\
     \textbf{}  &  \textbf{FVD($\downarrow$)}  & 13.25& 13.54 &13.57&13.45 & 13.69 & 13.17 & 12.12 &13.78&13.17&13.89&13.33 \\
\hline

        & \textbf{SSIM($\uparrow$)}  & 0.801& 0.834 & 0.805 &\textbf{0.813}& 0.804 & 0.793 & 0.828 &0.783&0.815&0.790&\textbf{0.802} \\

      \textbf{Full}  & \textbf{PSNR($\uparrow$)}  & 20.65& 21.40 & 20.34 &\textbf{20.79}& 20.44 & 20.48 & 20.97 &20.17&20.63&20.02&\textbf{20.45}\\

        & \textbf{FVD($\downarrow$)}  & 12.75& 11.25 & 12.93 &\textbf{12.31} & 11.01 & 11.04 & 11.76 &12.78&11.94&12.28&\textbf{11.80}\\

     \hline

    \end{tabular}
\end{table*}

\begin{table*}[]
 \small
  \centering
  \renewcommand{\tabcolsep}{0.6mm}  
  
  \caption{\small SSIM and PSNR scores for different combination of losses on NTU Dataset\cite{shahroudy2016ntu}  $56 \times 56$ resolution.  
  $L_2$ is the Reconstruction loss, $L_p$ is the Perceptual loss and $L_{adv}$ is the Adversarial loss. The left panel shows the results for multi-input view setup, whereas the right panel shows single-input view setup results.}
  \label{tab:loss_ablations}
    \begin{tabular}{|l|lcc|c|c|c|c||l|c|c|c|c|c|c|}
        \hline
         & \multicolumn{3}{c|} {\textbf{Losses}}  & \multicolumn{3}{c|} {\textbf{Multi-Input Setup}} &{\textbf{Avg}}&\multicolumn{6}{c|} {\textbf{Single-Input Setup}}& {\textbf{Avg}}  \\
        
        \cline{2-7}
        \cline{9-14}

        \textbf{\footnotesize{Metrics}}& \footnotesize{$L_2$}  & \footnotesize{$L_p$}& \footnotesize{$L_{adv}$} &  \textbf{($v_1 , v_2$) $\xrightarrow{}$ $v_3$} & \textbf{($v_2 , v_3$) $\xrightarrow{}$ $v_1$}  & \textbf{($v_1 , v_3$) $\xrightarrow{}$ $v_2$} &\footnotesize{\textbf{Score}} &  \footnotesize{$v_1$ $\xrightarrow{}$ $v_2$} & \footnotesize{$v_1$ $\xrightarrow{}$ $v_3$}  & \footnotesize{$v_2$ $\xrightarrow{}$ $v_1$} & \footnotesize{$v_2$ $\xrightarrow{}$ $v_3$} & \footnotesize{$v_3$ $\xrightarrow{}$ $v_1$} & \footnotesize{$v_3$ $\xrightarrow{}$ $v_2$} & \footnotesize{\textbf{Score}} 
        \\
        \hline
        \hline
         & \footnotesize{\checkmark}  &&&  0.699& 0.723 & 0.704 &0.708& {0.689} & 0.704 & 0.653 &0.678&0.651&0.670&0.674 \\

        &     &\footnotesize{\checkmark}&\footnotesize{\checkmark}& 0.737& 0.775 & 0.745 &0.752&0.703 & 0.714 & 0.601 &0.710&0.662&0.662&0.675 \\
  
       \textbf{\footnotesize{SSIM}} & \footnotesize{\checkmark}   &\footnotesize{\checkmark}&& 0.791& 0.833 & 0.796 &0.806&0.773 & 0.774 & 0.778 &0.768&0.766&0.749&0.768 \\
        
        \textbf{\footnotesize{($\uparrow$)}} & \footnotesize{\checkmark}   &&\footnotesize{\checkmark}&0.784 & 0.825 &0.789& 0.799& 0.783 & 0.776 & 0.798 &0.761&0.815&0.787& {0.786} \\
        & \footnotesize{\checkmark}   &\footnotesize{\checkmark}&\footnotesize{\checkmark}& 0.801& 0.834 & 0.805 &\textbf{0.813}&0.804 & 0.793 & 0.828 &0.783&0.815&0.790&\textbf{0.802} \\

        \hline
        & \footnotesize{\checkmark}  &&&17.56& 18.55 & 17.89 &18.00& 17.25 & 18.27 & 16.92 &17.27&16.55&17.01&17.21 \\

        &     &\footnotesize{\checkmark}&\footnotesize{\checkmark}& 18.96& 19.94 & 18.78 &19.22&17.85&   18.464 &15.82&18.39&17.34& 17.34&17.53 \\
  
       \textbf{\footnotesize{PSNR}} & \footnotesize{\checkmark}  &\footnotesize{\checkmark}&&  19.69& 21.63& 20.38 &20.72&19.76 & 20.24 & 19.72 &18.03&19.31&18.93&19.29 \\
        
        \textbf{\footnotesize{($\uparrow$)}}& \footnotesize{\checkmark}   &&\footnotesize{\checkmark}& 20.19& 21.06 & 19.91 &20.38&19.76 & 20.00 & 20.15 &18.58&19.55&18.8&19.47 \\
        
         & \footnotesize{\checkmark}  &\footnotesize{\checkmark}&\footnotesize{\checkmark}&20.65& 21.40 & 20.34 &\textbf{20.79}&20.44 & 20.48 & 20.97 &20.17&20.63&20.02&\textbf{20.45} \\

        \hline

    \end{tabular}
\end{table*}

\begin{figure}[]
\begin{center}
   \includegraphics[width=0.99\linewidth]{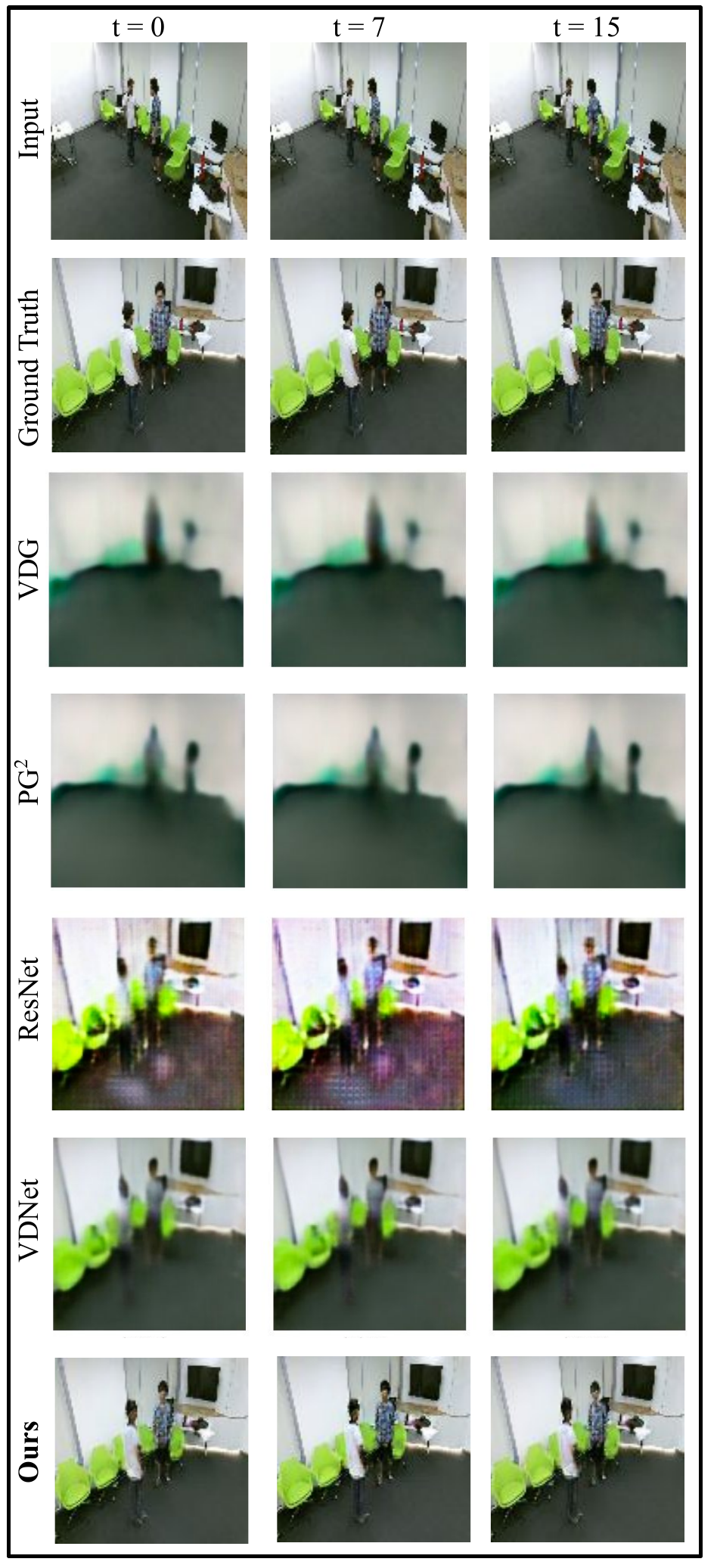}
\end{center}
   \caption{Qualitative results on NTU-RGB+D Dataset\cite{shahroudy2016ntu} showing comparison with VDG \cite{ilyes2018pose}, PG$^2$ \cite{ma2017pose}, ResNet \cite{he2015deep} and viewLSTM \cite{lakhal2019view}. We show two samples and the resolution is $112 \times 112$ with single-view input
  setup.}
\label{fig:ntu112_comparison_full}
\end{figure}

Secondly, we trained the network with only one representation at a time to investigate the impact of each branch in our architecture. 
To examine the role of View Dependent block, we employ View Dependent block $V_T$, in conjunction with Video Encoder $V_E$, and Video Decoder $V_D$ (i.e. Global Representation block $V_G$ is removed). To examine the role of Global Representation block, we employ Global Representation block $V_G$, in conjunction with Video Encoder $V_E$,  and Video Decoder $V_D$ (i.e. View Dependent block $V_T$ is removed).

The final experiment is performed on the full model $i.e.$ $V_T$ and $V_G$ combined. The qualitative results for network ablations are presented in Figure \ref{fig:network_ablations}. The first row shows the frames from input clip, followed by the synthesized outputs for View Dependent (VD) stream, Global Representation (GR) stream and the full model in rows two, three and four respectively. 
Finally, the ground truth frames are shown in the last row. 
The outputs from the network with View Dependent stream contains visible artifacts in the frames. The network with Global Representation stream shows improvements in visual quality of the output over the network with View Dependent branch only. Finally, the full model is able to add finer details like arm and leg (Figure \ref{fig:network_ablations}, right panel) in the synthesized output. 

Table \ref{tab:network_ablations_2to1} (Row 3-4) shows detailed quantitative results for network ablations for single and multiple input-view settings. The quantitative results are consistent for both single and multiple input-view setups, and for all input and output view combinations. The results show that View Dependent and Global Representation streams learn complementary features that are aggregated by the full model (Table \ref{tab:network_ablations_2to1} Row 5) and generate better results.


\begin{table}[t]
 \small
  \centering
\renewcommand{\tabcolsep}{3mm} 
  \caption{\small SSIM scores comparison with existing frame based methods\cite{ilyes2018pose, ma2017pose} and state-of-the-art video-based method, viewLSTM \cite{lakhal2019view} on \textbf{ NTU-RGB+D dataset at 112 $\times$ 112 resolution} with single-view input setup. 
  The scores for viewLSTM \cite{lakhal2019view} are computed for the videos synthesized using both depth and skeleton priors from the query view.}

  \label{tab:comparison_with_vLSTM_NTU}
    \begin{tabular}{|c|c|c|c|c|}
        \hline
      Metric  & \multicolumn{4}{c|}{Methods}
        \\
        \cline{2-5}
         &  {\cite{ilyes2018pose}} & {\cite{ma2017pose}}  & {\cite{lakhal2019view}} & \textbf{Ours} 
        \\
         \hline
         \hline
         \textbf{SSIM($\uparrow$)}&  {0.554} & {0.560}  & {0.783} & \textbf{0.791} 
        \\
        \hline
        


    \end{tabular}
\end{table}

\begin{figure}[t]
\begin{center}
  \includegraphics[width=.42\textwidth]{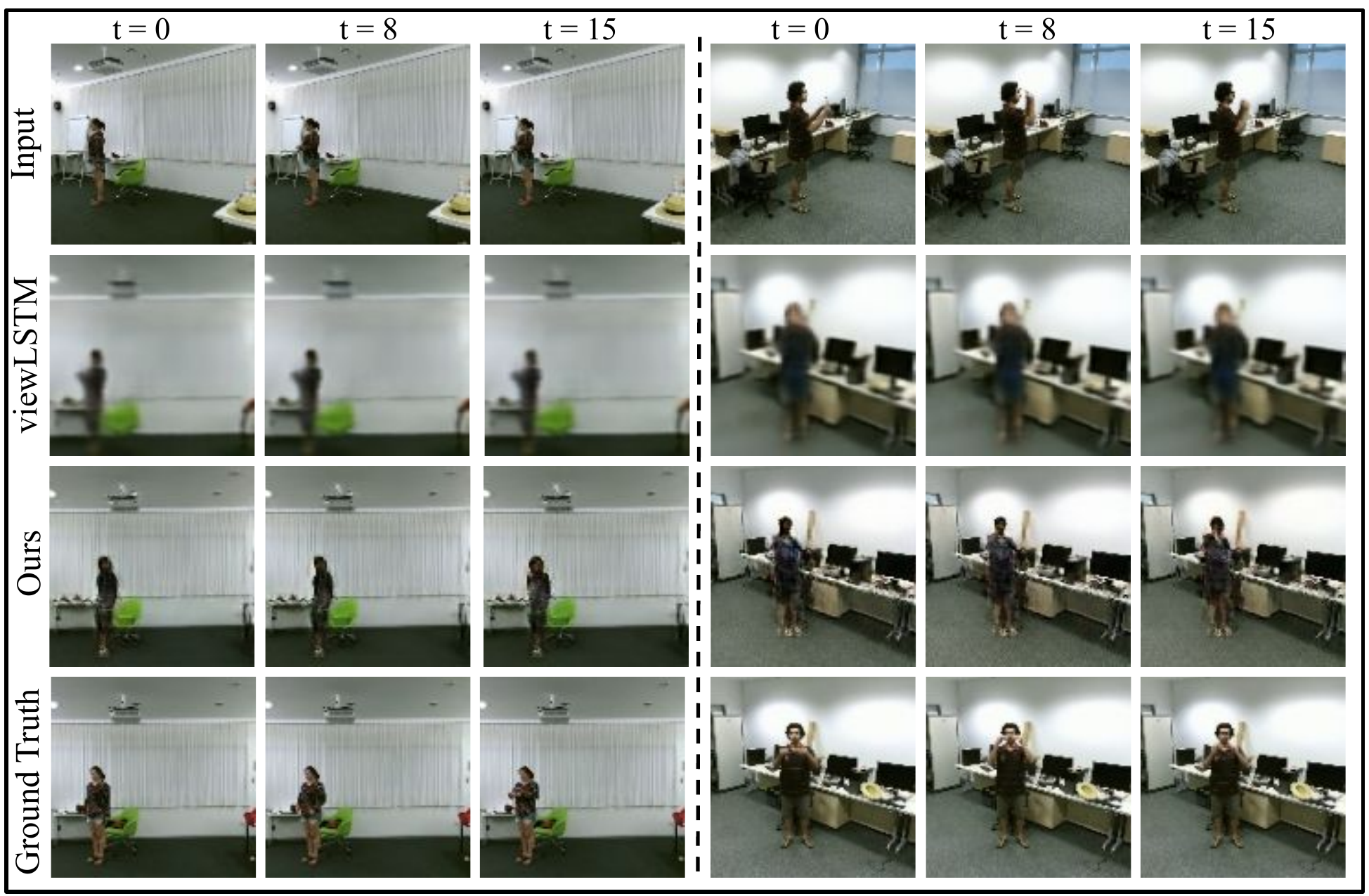}
\end{center}
  \caption{Qualitative results on NTU-RGB+D Dataset\cite{shahroudy2016ntu} showing comparison with viewLSTM \cite{lakhal2019view}. We show two samples and the resolution is $112 \times 112$ with single-view input
  setup.}
\label{fig:NTU112x112Comparison}
\end{figure}

\noindent \textbf{Comparison with Existing Works.}

In order to compare with the existing novel view prediction methods, 
we test our network to synthesize a query view video using a single view video as input.
We input multiple copies of the same input clip, as explained earlier with CMU Panoptic dataset. The experiment is conducted at 112 $\times$ 112 resolution.
We compare with a state-of-the-art video-based method, viewLSTM \cite{lakhal2019view} and two frame-based methods \cite{ilyes2018pose, ma2017pose}. 

The quantitative comparison is shown in Table \ref{tab:comparison_with_vLSTM_NTU}. It can be observed that SSIM with proposed approach is comparable to ViewLSTM \cite{lakhal2019view}, without using any priors from the query view. ViewLSTM \cite{lakhal2019view}, on the other hand, uses multiple priors from the query view, i.e both depth and skeleton information. The frame-based methods \cite{ilyes2018pose, ma2017pose} were reportedly trained to predict smaller view-point changes. When trained on NTU-RGB+D there are larger view point changes and our method predicts better quality videos. Also, these methods are unable to capture the temporal coherency between the synthesized frames.

\begin{table}[t]
 \small
  \centering
  \renewcommand{\tabcolsep}{2.0mm}  
  \caption{\small  SSIM, PSNR and FVD scores for different combination of losses on NTU-RGB+D Dataset \cite{shahroudy2016ntu} on $56 \times 56$ resolution. 
  }
  \label{tab:loss_ablations_2to11}
    \begin{tabular}{|ccc|c|c|c|}

\hline
  \multicolumn{3}{|c|}{Losses} & \multicolumn{3}{|c|}{Metrics}
\\
\hline
        \footnotesize{$L_2$}  & \footnotesize{$L_p$}& \footnotesize{$L_{adv}$} & \footnotesize{\textbf{SSIM($\uparrow$)}} & \footnotesize{\textbf{PSNR($\uparrow$)}}  & \footnotesize{\textbf{FVD($\downarrow$)}}   \\

        \hline
        \hline
        \footnotesize{\checkmark}  &&& 0.708& 18.00 & 15.42 \\

     & \footnotesize{\checkmark}&\footnotesize{\checkmark}& 0.752& 19.22 & 13.96\\
  
     \footnotesize{\checkmark}   &\footnotesize{\checkmark}&& 0.806& 20.72 & 13.28  \\
        
      \footnotesize{\checkmark}   &&\footnotesize{\checkmark}& 0.799 & 20.38 &13.76 \\
        \footnotesize{\checkmark}   &\footnotesize{\checkmark}&\footnotesize{\checkmark}& \textbf{0.817}& \textbf{20.77} & \textbf{12.31}  \\

        \hline

    \end{tabular}

\end{table}

Figure \ref{fig:NTU112x112Comparison} shows a qualitative comparison between the outputs synthesized by viewLSTM \cite{lakhal2019view} (second row) and our proposed method (third row) on NTU-RGB+D dataset. The samples are shown for three time steps and for two samples. As seen with the SSIM scores, our method produces sharper foreground objects and better background details.

We provide more qualitative results comparing with our baselines VGD \cite{ilyes2018pose}, PG$^2$ \cite{ma2017pose}, ResNet \cite{he2015deep} and viewLSTM \cite{lakhal2019view} in Figure \ref{fig:ntu112_comparison_full}. As can be seen, the frame-bases baselines VDG and PG$^2$ are not able to produce even the scene structure correctly. ResNet, which is used as a video-based baseline, generates correct scene structure but has visible artifacts present. VDNet produces better results but the video quality is low. Finally, we can observe that the proposed method perfectly generates the background as well as the actor performing the action with better motion as well.

\begin{figure}[]
\begin{center}
   \includegraphics[width=0.89\linewidth]{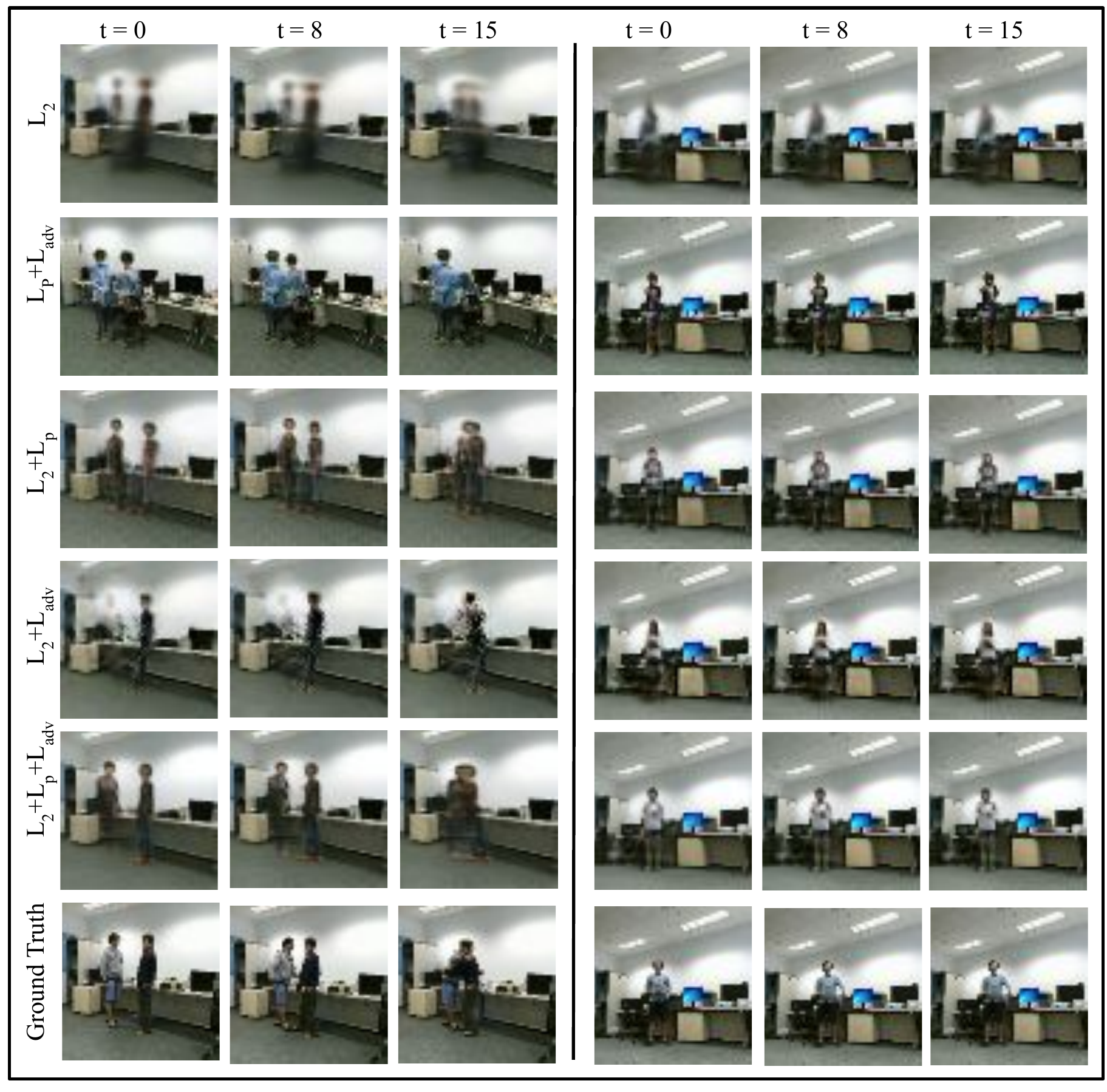}
\end{center}
   \caption{\small Qualitative Results for loss ablations on NTU Dataset; We show frames obtained by employing  different losses.
   We obtained the the best visual results  employing  all three losses ($L_2 + L_p + L_{adv}$);  here $L_2$ : Reconstruction Loss,  $L_p$ : Perceptual Loss, $L_{adv}$ : Adversarial Loss.
   }
\label{fig:loss_ablations}
\end{figure}

\noindent \textbf{Loss Ablations. }
Figure \ref{fig:loss_ablations} shows the qualitative results obtained by employing different  losses. 
 Using only the reconstruction loss produces very blurry results. Adversarial and perceptual loss combined ($L_p + L_{adv}$) helps in sharpening the frames but does not get the color information and background correctly. Adding perceptual loss with reconstruction loss ($L_2 + L_p$) gets the background information well, but the structures of bodies are not defined. $L_2 + L_{adv}$ also produces sharp results with correct background but lacks in producing correct body structures. Finally, combining $L_2 + L_p + L_{adv}$, we get improved results compared to other losses. 
 
 Table \ref{tab:loss_ablations} shows detailed quantitative results for loss ablations for single input-view settings. The results are consistent for different input and output view combinations. The use of all three losses helps to synthesize query view outputs better than using only one or two loss combinations as validated by the table.

\end{document}